\documentclass[10pt,letterpaper]{article}

\usepackage{cogsci}
\usepackage{pslatex}
\usepackage{apacite}
\usepackage{amsmath}
\usepackage{dsfont}
\usepackage{caption}
\usepackage{subcaption}
\usepackage{graphicx}
\usepackage{algorithm} 
\usepackage{algpseudocode} 
\usepackage{float}
\cogscifinalcopy

\title{What are the mechanisms underlying metacognitive learning?}
\author{{\large \bf Ruiqi He (ruiqi.he@tuebingen.mpg.de)} \\
  Max Planck Institute for Intelligent Systems\\
  Stuttgart, Germany
  \AND {\large \bf Falk Lieder (falk.lieder@tuebingen.mpg.de)} \\
    Max Planck Institute for Intelligent Systems\\
  Stuttgart, Germany}

\begin{document}

\maketitle

\begin{abstract}
How is it that humans can solve complex planning tasks so efficiently despite limited cognitive resources? One reason is its ability to know how to use its limited computational resources to make clever choices. 
We postulate that people learn this ability from trial and error (\textit{metacognitive reinforcement learning}). Here, we systematize models of the underlying learning mechanisms and enhance them with more sophisticated additional mechanisms. We fit the resulting 86 models to human data collected in previous experiments where different phenomena of metacognitive learning were demonstrated and performed Bayesian model selection. Our results suggest that a gradient ascent through the space of cognitive strategies can explain most of the observed qualitative phenomena, and is therefore a promising candidate for explaining the mechanism underlying metacognitive learning.  

\textbf{Keywords:} 
metacognitive learning, planning, strategy discovery, cognitive modelling, reinforcement learning
\end{abstract}

\section{Introduction}
Humans frequently face complex problems that requires planning long chains of actions to accomplish far-off objectives. A search tree can represent the space of potential future actions and outcomes, which expands exponentially as the length of the sequences increases. While exponential growth in computational power enables current trends in artificial intelligence, the cognitive capabilities of the human mind are much more constrained.
Therefore, people have to make efficient use of their limited cognitive resources (\textit{resource-rationality}) \cite{lieder2020resource}. So, how is it possible that people can still plan so efficiently? One potential explanation is that meta-reasoning, the ability to reason about reasoning, might help people to accomplish more with less computational effort \cite{Griffiths2019}. 
In the context of planning, this means making wise choices about when and how to plan, that is whether and how to use computational resources. 
However, according to \citeA{Russell1991}, optimal meta-reasoning is often regarded as an intractable problem. This raises the question of how people can nonetheless solve the intractable meta-reasoning problem. One possibility is that people learn an approximate solution via trial and error, an idea known as \textit{metacognitive reinforcement learning} \cite{LiederGriffiths2017,krueger2017enhancing, LiederShenhav2018}.

This idea has been used in earlier research to explain how people learn to select between various cognitive strategies \cite{Erev2005,rieskamp2006ssl,LiederGriffiths2017}, how many steps to plan ahead \cite{krueger2017enhancing} and when to exercise how much cognitive control \cite{LiederShenhav2018}. In the context of planning, previous work suggests that metacognitive reinforcement learning adapts which information people prioritise in their decisions \cite{jain2019measuring,HeJainLieder2021} and how much planning they perform to the costs and benefits of planning \cite{HeJainLieder2021NIPS-Planning}. While previous work each focused on explaining individual aspects of metacognitive learning with a small set of models, none of the models was tested to explain \textit{all} observed qualitative phenomena. In addition, previous findings paint a rather inconsistent and even contradictory picture of how people learn planning strategies, with different articles arguing for different learning mechanisms \cite{Jain2019CCN, HeJainLieder2021NIPS-Planning}.

Therefore, in this work, we investigate whether there is one single metacognitive reinforcement learning model that can largely explain all observed phenomena. Our contribution is two-fold: i) We systematically compare all existing models on data collected in empirical experiments where learning-induced changes in people's planning strategies were demonstrated, and ii) we extend existing models to systematically formalize plausible alternative assumptions and all of their possible combinations. This led to 86 different models, which we fit using maximum likelihood criterion and compare using Bayesian model selection, as well as perform model simulation. The winning model gives us an indication of the underlying mechanisms of how people learn planning strategies. 

This line of research contributes to the larger goal of understanding metacognitive learning. It also provides a foundation for training programs aiming to improve human decision-making and to help people overcome maladaptive ways of learning planning strategies. 

\section{Background}
To model the mechanism of metacognitive learning, we take inspiration from reinforcement learning algorithms and use the framework of meta-decision-making, which we will now briefly introduce and explain how they can be combined into a framework called \textit{metacognitive reinforcement learning}. 

\subsection{Reinforcement learning}
Previous studies suggest that human learning is motivated by reward and penalties gained through trial and error \cite{niv2009reinforcement}, which builds the foundation of reinforcement learning algorithms that learn to predict the potential reward from performing a specific action $a$ in a specific state $s$. This estimate $Q(s,a)$ is updated according to the reward prediction error $\delta$, which is the difference between actual and expected rewards:
\begin{equation} \label{eq:QLearning}
    Q(s,a) \leftarrow Q(s,a) - \alpha \cdot \delta 
\end{equation}
where $Q$ denotes the Q-value \cite{watkins1992q} and $\alpha$ is the learning rate. 
To compromise between exploitation and exploration, the agent can pick its actions \textit{probabilistically}, maximising the predicted action value, for example using the softmax rule \cite{Williams1992} $P(a|s,Q) \propto \exp( 1/ \tau \cdot Q(s,a))$ 
% \begin{equation}
%     P(a|s,Q) \propto \exp( 1/ \tau \cdot Q(s,a))
% \end{equation}
where $\tau$ is the inverse temperature parameter. 

\subsection{Meta-decision-making}
The brain is supposedly equipped with multiple decision systems that interact in various ways \cite{Dolan2013,Daw2018}. The model-based system, in contrast to Pavlovian and model-free systems, allows for flexible reasoning about which action is preferable but demands a process for deciding which information should be considered for a given decision. Therefore, an important part of deciding how to decide is to efficiently balance decision quality and decision time, known as \emph{meta-decision-making} \cite{Boureau2015}. 
The problem of meta-decision-making has been recently formalized as a meta-level MDP \cite{krueger2017enhancing,Griffiths2019}:
\begin{equation}
    M_{meta}=\left( \mathcal{B}, \mathcal{C} \cup \lbrace \bot \rbrace, T_{meta}, r_{meta} \right),\label{eq:metaLevelMDP}
\end{equation}
where belief states $b_t \in \mathcal{B}$ denotes the model-based decision system's beliefs about the values of actions. 
The computations of the decision system ($c_1,c_2,\cdots$) probabilistically determine the temporal development of those belief states $b_1,b_2,\cdots$ according to the meta-level transition probabilities $T_{\text{meta}}(b_t,c_t,b_{t+1})$. 
The meta-level reward function $r_{\text{meta}}(b_t,c_t)$ encodes the cost of performing the planning operation $c_t\in \mathcal{C}$ and the expected return of terminating planning ($c_t=\bot$) and acting based on the current belief state $b_t$. Reinforcement learning algorithms, such as Q-learning (see Equation~\ref{eq:QLearning}), can be used to solve this meta-level MDP.

\subsection{Metacognitive reinforcement learning}
Finding efficient planning strategies can be formalized as solving a metalevel MDP for the best metalevel policy \cite{Griffiths2019}. 
However, as it is often computationally intractable to solve meta-decision-making problems optimally, we will assume that the brain approximates optimal meta-decision-making through reinforcement learning mechanisms \cite{Russell1991,callaway2018learning} that attempt to approximate the optimal solution of the meta-level MDP defined in Equation~\ref{eq:metaLevelMDP} by either learning to approximate the optimal policy directly \cite{HeJainLieder2021} or by learning an approximation to its value function \cite{jain2019measuring}.

\subsection{Experiments}
\begin{figure}
    \centering
\includegraphics[width=0.5\linewidth, height=4.5cm]{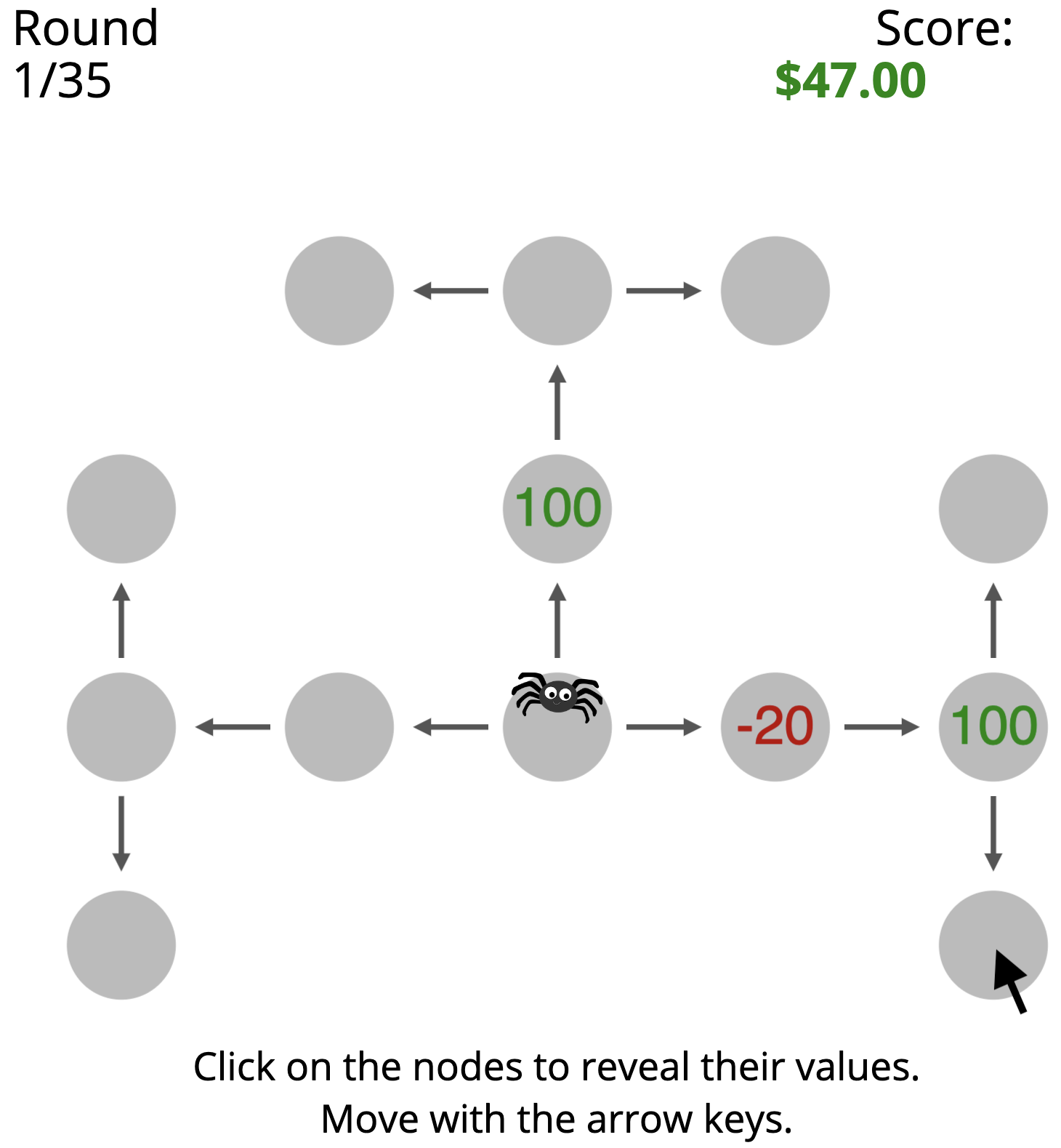}
    \caption{Exemplary trial of the planning task}
    \label{fig:webofcash}
\end{figure}

For testing the ability of our models to explain different aspects of metacognitive learning, we will use data from previous work that examined several aspects of metacognitive learning in the domain of planning. \citeA{HeJainLieder2021} and \citeA{HeJainLieder2021NIPS-Planning} utilized the Mouselab-MDP paradigm to design two experiments where participants were asked to perform repeated trials of a planning task (see Figure~\ref{fig:webofcash}). The goal in the experiment was to collect a high score, which signals the adaptiveness and resource-rationality \cite{lieder2020resource} of the participant at a given trial. The rewards are initially hidden but can be revealed by clicking on the nodes. Each click has a cost. Participants' clicks were recorded because they indicate planning operations that people perform to estimate the values of alternative future locations. 

\subsubsection{Adaptation to different environment structures}
In the first experiment, participants were randomly allocated to one of three conditions, where the environment structure rendered either long-term planning (examining the farthest nodes), short-term planning (examining immediate nodes) or best-first search planning (starting with examining immediate and middle nodes and continue to examine other nodes according to the most promising ones) most beneficial. The results suggested that people gradually learn to use the corresponding adaptive strategies for each environment. 

\subsubsection{Adaptation of the amount of planning depending on the costs and benefits of planning}
The second experiment indicated that people do learn how much to plan. For this, participants were assigned to one of four different conditions, each differed in the benefit and cost of planning. Their number of clicks indicated whether participants learned to adapt their amount of planning depending on the condition. 

\section{Models and methods}
The models of metacognitive learning we test in this article have three components: i) the representation of the planning strategies that the learning mechanism operates on, ii) the basic learning mechanism, and iii) additional attributes. The following three sections introduce these components. We then describe how the models were fit and selected. 
%Therefore, we will first introduce previous work on strategy representation, explain three existing learning mechanisms of metacognitive learning as well as one non-learning alternative and then explain promising extensions to the models followed by elaborating on our model fitting and selection procedure. %Combining the basic learning mechanisms with the model attributes results in 86 different models. A list of the models can be found in \url{https://osf.io/wz9uj/}. 

\subsection{Mental representation of planning strategies}
The planning strategies are modelled as softmax policies that depend on a weighted combination of 56 features \cite{Jain2021Computational}. %The features are designed to represent the planning strategies people use in the corresponding task at a given trial, and they are also informed by previous work on the brain's decision systems. For a detailed description of all features, see \citeA{Jain2021Computational} 
For instance, one group of features is related to pruning \cite{Huys2012}, which is associated with giving a negative value to consider a path whose predicted value is below a specific threshold. Therefore, using this representation, a person's learning trajectory can be described as a time series of the weight vectors that correspond to their planning strategies in terms of those features.
%The features are designed so that they can represent the planning strategies people have been found to use in this task, and they are also informed by previous work on the brain's decision systems. 
%For instance, one group of features is related to pruning \cite{Huys2012}, which are associated with giving a negative value to considering a path whose predicted value is below a specific threshold. % For a detailed description of all features, see \citeA{Jain2021Computational}.

\subsection{Basic learning mechanisms}

We consider four possible basic learning mechanisms: learning the value of computation, gradient ascent through the strategy space, forming a mental habit, and no learning.

\subsubsection{Learning the value of computation}
According to the Learned Value of Computation (LVOC) model, people learn how valuable it is to perform each planning operation depending on what is already known \cite{krueger2017enhancing}. This is achieved by approximating the meta-level Q-function by a linear combination of the features mentioned above:
\begin{equation}
	Q_{\text{meta}}(b_k,c_k) \approx \sum_{j=1}^{56} w_j \cdot f_j(b_k,c_k),\label{eq:FunctionApproximation}
\end{equation}
The weights of those features are learned by Bayesian linear regression of the bootstrap estimate $\hat{Q}(b_k,c_k)=r_{\text{meta}}(b_k,c_k)+ \langle \mu_t , \mathbf{f}(b',c') \rangle \label{eq:bootstrap}$
%$\hat{Q}(b_k,c_k)$ of the meta-level value function onto the features $\mathbf{f}$:
% \begin{equation}
% 	\hat{Q}(b_k,c_k)=r_{\text{meta}}(b_k,c_k)+ \langle \mu_t , \mathbf{f}(b',c') \rangle \label{eq:bootstrap}
% \end{equation}
which is the sum of the immediate meta-level reward and the anticipated value of the future belief state $b'$ under the present meta-level policy. The predicted value of $b'$ is the scalar product of the posterior mean $\mu_t$ of the weights $\mathbf{w}$ given the observations from all preceding planning operations and the features $\mathbf{f}(b',c')$ of $b'$ and the cognitive operation $c'$ that the current policy picks given state. 
To make the $k$\textsuperscript{th} planning operation, $n$ weight vectors are sampled from the posterior distribution using a generalized Thompson sampling $\tilde{w}^{(1)}_k, \cdots, \tilde{w}^{(n)}_k \sim P(\mathbf{w} | \mathcal{E}_k)$, 
% \begin{equation}
% 	\tilde{w}^{(1)}_k, \cdots, \tilde{w}^{(n)}_k \sim P(\mathbf{w} | \mathcal{E}_k),
% \end{equation}
where the set $\mathcal{E}_k=\lbrace e_1, \cdots, e_k \rbrace$ contains the meta-decision-maker's experience from the first $k$ meta-decisions. Each meta-level experience $e_i\in\mathcal{E}_k$ is a tuple $\left(b_i,h_i,\hat{Q}(b_i,c_i; \mu_i)\right)$ containing a meta-level state, the selected planning operation in it, and the bootstrap estimates of its Q-value. The arithmetic mean of the sampled $n$ weight vectors is then used to predict the Q-values of each potential planning operation $c \in \mathcal{C}$ according to Equation~\ref{eq:FunctionApproximation}. 
The LVOC model therefore has the following free parameters: $p$, the mean vector $\mu_{prior}$ and variance $\sigma^2_{\text{prior}}$ of its prior distribution $\mathcal{N}(\mathbf{w};\mu_{prior}, \sigma^2\cdot\text{I} )$ on the weights $\mathbf{w}$, and the number of samples $n$.

\subsubsection{Gradient ascent through the strategy space}
According to the REINFORCE model \cite{jain2019measuring}, which is based on the REINFORCE algorithm \cite{Williams1992}, metacognitive learning proceeds by gradient ascent through the space of possible planning strategies. When a plan is executed and its outcomes are observed, the weights $w$ representing the strategy are adjusted in the direction of the gradient of the return\footnote{The return is the  the sum of the rewards along the chosen path minus the cost of the performed planning operations.}, that is
\begin{equation}
    \mathbf{w} \leftarrow \mathbf{w} + \alpha \cdot \sum_{t=1}^{O}\gamma^{t-1} \cdot r_{meta}(b_t,c_t) \cdot \nabla_\mathbf{w} \ln \pi_\mathbf{w}(c_t | b_t), \label{eq:reinforceUpdate}
\end{equation}
where $\gamma$ is the discount factor, and $O$ is the number of planning operations executed by the model on that trial. The learning rate $\alpha$ is optimised using ADAM \cite{kingma2014adam}. %Next to the initial weight vector $\mathbf{w}$, 
The REINFORCE model has three free parameters: $\alpha$, $\gamma$ and inverse temperature $\tau$ that are fit separately for each participant. The weights are initialised randomly. 

% The likelihood of an action is obtained through the policy:  
% \begin{equation}
%    \pi_\theta(c | b) = \frac{\exp\left( \frac{1}{\tau} \cdot \sum_{k=1}^{52} \theta_k \cdot f_k(b,c) \right)}{\sum_{c \in \mathcal{C}_{b}} \exp\left( \frac{1}{\tau} \cdot \sum_{k=1}^{52} \theta_k \cdot f_k(b,c) \right)}
%    \label{eq:softmaxreinforce}
% \end{equation}
% where $b$ represents the belief state, $c$ represents the click under consideration and $\mathcal{C}_{b}$ is the set of clicks available in belief state $b$. 

\subsubsection{Mental habit formation}
This model assumes that the only mechanism through which people's planning strategies change is the formation of mental habits. Following \citeA{miller2019habits} and \citeA{morris2022habits}, this model assumes that people's propensity to perform a given (type of) planning operation increases with the number of times they have performed it in the past. This is implemented as a softmax decision rule applied to a weighted sum of frequency-based features, including the number of previous clicks on the same node, the same branch, and the same level, respectively. %For a detailed description of all of these features, see REF.
%nodes on the same branch as the clicked node and how many nodes on the same level as the clicked node has been observed before.
%does not perform any parameter updates but uses habit-related features that describe the total numbers of clicks, how many nodes on the same branch as the clicked node and how many nodes on the same level as the clicked node has been observed before. 

\subsubsection{Non-learning model}
This model does not perform any parameter updates and does not use habitual features. 

\subsection{Extensions}
We augmented the REINFORCE and LVOC models with three optional components: a two-stage hierarchical meta-decision-making process (\textit{hierarchical meta-control}), metacognitive rewards for generating valuable information (\textit{pseudo-rewards}), and deliberating about the value of termination when taking an action (\textit{termination deliberation}).  

\subsubsection{Hierarchical meta-control}
Previous research suggests that foraging decisions are made by two distinct decision systems: the ventromedial prefrontal cortex and the dorsal anterior cingulate cortex \cite{Rushworth2012}. We, therefore, developed an extension that first decides whether to continue planning (Stage 1) and then selects the next planning operation according to either the LVOC or the REINFORCE model (Stage~2) if it chose to continue planning in Stage~1. 
For Stage~1, our models consider three potential decision rules. Each decision rule %adaptive satisficing, confidence bound and threshold. The probability of termination is determined using 
is a tempered sigmoid function $\sigma(x, \tau) = (1+e^{- \frac{x}{\tau}})^{-1}$ \cite{papernot2021tempered}.
In each case, the function's argument $x$ is a different function $f(\mathds{M})$ of the expected sum of rewards along the best path according to the information observed so far ($\mathds{M} = \max_{path}\mathds{E}[R(\text{path}) \mid b]$). Concretely, the three stopping rules compare $\mathds{M}$ against a fixed threshold, a threshold that tracks the outcomes of previous trials, and a
threshold that decreases with the number of clicks, respectively.  %If no nodes are revealed, $\mathds{M}$ will equal 0. %The three stopping rules differ in the function $f$.

% \begin{equation} \label{eq:M}
%     \mathds{M} = \max_{path}\mathds{E}[R(\text{path}) \mid b]
% \end{equation}

\paragraph{Fixed threshold}
This decision rule probabilistically terminates planning when the normalized value of $\mathds{M}$ reaches the threshold $\eta$, that is
$  P(C = \bot \mid b) = \sigma\left(\frac{\mathds{M} - v_{\text{min}}}{v_{\text{max}} - v_{\text{min}}} - \eta, \tau \right)$, where $v_{\text{min}}$ and  $v_{\text{max}}$ are the trial's lowest and highest possible returns, respectively. 

\paragraph{Decreasing threshold}
Building on the observation that the threshold of the resource-rational planning strategy decreases with the number of clicks \cite{callaway2018resource}, this decision rule adjusts the threshold based on the number of clicks made so far ($n_c$), that is:
$ P(C = \bot \mid b) = \sigma(\mathds{M} - e^a + e^b \cdot n_c, \tau)$, 
%\sigma(\mathds{M} - e^{a} + e^{b} \cdot n_c, \tau) 
where $a$ and $b$ are the free parameters. 

\paragraph{Threshold based on past performance}
This decision rule models the idea that people learn what is good enough from experience. Concretely, this decision rule assumes that the threshold  $M\sim \mathcal{N}\left(m; \frac{\eta}{\sqrt{n + 1}}\right)$ is a noisy estimate of the average $m$ of their previous scores, that is
$P(C = \bot \mid b) = \sigma(\mathds{M}- M), \tau)$, 
where $n$ is the number of trials and $\eta$ is a free parameter. The probability distribution of the threshold is derived from the assumption that the threshold is an average of noisy memories of previous scores. 

\subsubsection{Pseudo-rewards}
The central role of reward prediction errors in reinforcement learning \cite{Schultz1997,Glimcher2011} and the dearth of external reward in metacognitive learning \cite{Hay2016} indicate that the brain might accelerate the learning process by producing additional metacognitive pseudo-rewards that convey the value of information produced by the last planning operation. Concretely, the pseudo-reward (PR) for transitioning from belief state $b_t$ to $b_{t+1}$ is the difference between the expected value of the path that the agent would have taken in the previous belief state $b_{t}$ and the expected value of the best path in the new belief state $b_{t+1}$: $\text{PR}(b_t,c,b_{t+1})=\mathds{E}[R_{\pi_{b_{t+1}}}|b_{t+1}]-\mathds{E}[R_{\pi_{b_t}}|b_{t+1}]$ where $\pi_b(s) = \text{\text{argmax}}_a \mathds{E}_b[R \mid s, a]$ is the policy the agent will use to navigate the physical environment when its belief state is $b$, and $R$ is the expected value of the sum of the external rewards (e.g., the sum of rewards collected by moving through the planning task) according to the probability distribution $b$.
% \begin{equation}
%     \text{PR}(b_t,c,b_{t+1})=\mathds{E}[R_{\pi_{b_{t+1}}}|b_{t+1}]-\mathds{E}[R_{\pi_{b_t}}|b_{t+1}], \label{eq:PR}
% \end{equation}
%where $\pi_b(s) = \text{\text{argmax}}_a \mathds{E}_b[R \mid s, a]$ is the policy the agent will use to navigate the physical environment when its belief state is $b$, and $R$ is the expected value of the sum of the external rewards (e.g., the sum of rewards collected by moving through the maze) according to the probability distribution $b$.

\subsubsection{Termination deliberation}
If people engaged in rational metareasoning \cite{Griffiths2019}, they would calculate the expected value of acting on their current belief $b$ from the information it encodes (\textit{termination deliberation}). Alternatively, people might learn when to terminate through the same learning mechanism through which they learn to select between alternative planning operations (no termination deliberation). %We, therefore, developed two versions of each model: one with termination deliberation and one without. 

\subsection{Model fitting}
Combining the basic learning mechanisms with the model attributes resulted in 86 different models (see \url{https://osf.io/wz9uj/} for a list of all model). 
We fitted all models to 382 participants from both experiments by maximizing the likelihood function of the participants' click sequences using Bayesian optimization \cite{bergstra2013making}. The likelihood of a click sequence is the product of the likelihood of the individual clicks. 

% \begin{algorithm}
% 	\caption{Model fitting procedure} 
% 	\begin{algorithmic}[1]
%     \State Initiate free parameters
% 		\For {400 iterations}
%             \State For REINFORCE: initiate features weights $\theta$
%             \State For LVOC: initiate $\mu$ and $\Sigma$
% 			\For {35 trials}
%                 \For{$c$ clicks}
% 				\State Calculate likelihood of a click (Equation~\ref{eq:likelihood})
% 				\State Obtain reward
%                 \State For LVOC: Update $\mu$ and $\Sigma$ using rewards
%             \EndFor
%             \State For REINFORCE: Update feature weights $\theta$ using Equation~\ref{eq:reinforceUpdate} using rewards
% 			\EndFor
% 		\State Optimize parameters using Bayesian optimisation 
% 		\EndFor
% 	\end{algorithmic} 
%  \label{algorithm}
% \end{algorithm}
%To compromise between exploitation and exploration, the model pick its clicks \textit{probabilistically} by using the softmax rule \cite{Williams1992}. Therefore, the likelihood of a certain click $c$ is: 
%\begin{equation}
%    P(c|b,w) = \frac{\exp\left( \frac{1}{\tau} \cdot \sum_{k=1}^{56} w_j \cdot f_k(b,c) \right)}{\sum_{c \in \mathcal{C}_{b}} \exp\left( \frac{1}{\tau} \cdot \sum_{k=1}^{56} w_j \cdot f_k(b,c) \right)} \label{eq:likelihood}
% \end{equation}
% where \textit{w} is updated differently depending on the corresponding model. The likelihood of a click sequence is the product of the likelihood of the individual clicks. 

\subsection{Model selection}
After having fitted the models, we perform model selection using the Bayesian information criterion (BIC) \cite{schwarz1978estimating}, and Bayesian model selection (BMS). Concretely, we estimate the expected proportion of people who are best described by a given model ($r$) and the \textit{exceedance} probability $\phi$ that this proportion is significantly higher than the corresponding proportion for any other model by using random effect Bayesian model selection \cite{rigoux2014bayesian,stephan2009bayesian}. To obtain the equivalent conclusions for groups of models that share some feature, we perform family-level Bayesian model selection \cite{penny2010comparing}.

\section{Results}
The code and the BIC of all 86 models can be found in \url{https://osf.io/wz9uj/}. 

\subsection{Comparing all models for all participants}
To examine which of the learning mechanisms can best explain human behavior, we grouped the models into 4 model families: non-learning, mental habit, LVOC, and REINFORCE models. We found that the model family whose members provided the best explanation for the largest number of participants was the REINFORCE models (see Table~\ref{table:familybmsall}), which explained about 41.13\% of the participants better than models from other model families. 
\begin{table}[h!]
\small
\centering
\parbox{.45\linewidth}{
\begin{tabular}{ccc}
Model family & $r$ & $\phi$ \\ \hline
Non-learning & 0.34 & 0.06 \\
Mental habit & 0.06 & 0 \\
LVOC & 0.20 & 0 \\
REINFORCE & 0.41 & 0.94 \\ \hline
\end{tabular}
\caption{Family-level BMS for all participants}
\label{table:familybmsall}
}
\hfill
\parbox{.45\linewidth}{
\centering
\begin{tabular}{ccc}
Model family & $r$ & $\phi$ \\ \hline
Non-learning & 0.25 & 0 \\
Mental habit & 0.07 & 0 \\
LVOC & 0.21 & 0 \\
REINFORCE & 0.47 & 1 \\ \hline
\end{tabular}
\caption{Family-level BMS for learners}
\label{table:familybmslearner}
}
\end{table}
The second most successful model family is the non-learning model. It provided the best explanation for 34\% of the participants, which is mainly driven by the similarly high proportion of participants who did not show any signs of learning. Therefore, for the remaining analysis, we will focus on participants who demonstrated learning. That is, participants who changed their planning strategies at least once in the first experiment and participants whose planning amount changed significantly during the second experiment (determined using the Mann-Kendall test of trend; all $S>105$ for increasing trend, all $S<-57$ for decreasing trend, all $p<.05$). This led to the selection of 224 participants (58.64\%). 

\subsection{Comparing all models for learners}
Family-level BMS for the remaining participants showed a decrease in the proportion of participants best explained by the non-learning model to 25\%, while the proportion of participants best explained by the learning models increased to 75\%. REINFORCE models now explain the data from 47.41\% of the learners better than the other models (see Table~\ref{table:familybmslearner}).
\begin{table}[h!]
\small
\centering
\begin{tabular}{cccc}
Model & $r$ & $\phi$  & BIC\\ \hline
REINFORCE with PR & 0.11 & 0.80  & 569.09\\
Plain REINFORCE & 0.08 & 0.15 & 567.99\\
LVOC & 0.07 & 0.03 &  589.76\\
Mental-habit & 0.06 & 0.01 & 598.90\\\hline
\end{tabular}
\caption{Model-level BMS and BIC of the learning models for all learners across both experiments.}
\label{table:reinforce}
\end{table}
Comparing the learning models individually, the models that were best for the highest proportion of participants are the REINFORCE model with pseudo-reward and the plain REINFORCE model, followed by the plain LVOC model, and the mental-habit model (see Table~\ref{table:reinforce}).
\begin{table}
\small
\centering
\parbox{.45\linewidth}{
\begin{tabular}{ccc}
Model family & $r$ & $\phi$ \\ \hline
Mental habit & 0.09 & 0 \\
LVOC & 0.19 & 0 \\
REINFORCE & 0.72 & 1 \\ \hline
\end{tabular}
\caption{Family-level BMS for Experiment 1}
\label{table:exp1exp2_family1}
}
\hfill
\parbox{.45\linewidth}{
\centering
\begin{tabular}{ccc}
Model family & $r$ & $\phi$ \\ \hline
Mental habit & 0.10 & 0 \\
LVOC & 0.40 & 0.17 \\
REINFORCE & 0.50 & 0.83 \\ \hline
\end{tabular}
\caption{Family-level BMS for Experiment 2}
\label{table:exp1exp2_family2}
}
\end{table}
\begin{table}[h!]
\small
\centering
\begin{tabular}{ccccc}
Exp & Model & $r$ & $\phi$ & BIC\\ \hline
1 & REINFORCE with PR & 0.11 & 0.88 & 600.39 \\
1 & Plain REINFORCE & 0.07 & 0.07 & 604.15 \\ \hline
2 & REINFORCE with PR & 0.06 & 0.11 & 510.51\\
2 & Plain REINFORCE & 0.07 & 0.26 & 500.29\\ \hline
\end{tabular}
\caption{Model-level BMS and BIC of the models with largest $r$ for Experiments 1 and 2.}
\label{table:exp1exp2}
\end{table}

To examine whether how people learn adaptive planning strategies (Experiment~1) and how people adapt their amount of planning (Experiment~2) can be largely explained by a single model, we combined the data from both experiments into one data set. For both experiments, the REINFORCE learning mechanism explains the largest proportion of participants (see Table~\ref{table:exp1exp2_family1} and \ref{table:exp1exp2_family2}). While a larger proportion of participants are best explained by the REINFORCE model with pseudo-reward in Experiment 1, Experiment 2 favors the plain REINFORCE model (see Table~\ref{table:exp1exp2}). 
BMS family-level comparison on models with pseudo-reward against models without pseudo-reward on both experiments combined yielded $r = 0.42, \phi = 0.01$ for models with pseudo-reward and $r = 0.58, \phi = 0.99$ for models without. 
%To further examine the attribute of pseudo-reward, we performed family-level BMS with two model-families: one with pseudo-rewards and one without them (see Table~\ref{table:prbms}). 
% \begin{table}[h!]
% \small
% \centering
% \begin{tabular}{ccc}
% Model family & $r$ & $\phi$ \\ \hline
% Pseudo-reward & 0.42 & 0.01\\
% No pseudo-reward & 0.58 & 0.99 \\ \hline
% \end{tabular}
% \caption{Family-level BMS comparing models with and without pseudo-reward for all learners for both experiments}
% \label{table:prbms}
% \end{table}
Comparing the difference in BIC between the REINFORCE model with pseudo-rewards and its plain version for all learners revealed substantial evidence for the absence of pseudo-rewards in 104 out of 224 participants ($\Delta$BIC $>3.2$ for 65 participants of Exp.~1 and 39 from Exp.~2) and substantial evidence for its presence in 104 other participants ($\Delta$BIC$<-3.2$ for 75 participants of Exp.~1 and 29 from Exp.~2). The remaining 16 participants' absolute difference in BIC was less than 3.2. This suggests that about half of the participants (42\%) might use intrinsically generated pseudo-rewards to inform the metacognitive learning, while the other half (58\%) do not. 
A $\chi^2$ test comparing the proportion of participants whose data is better explained by a model with pseudo-rewards between the two experiments yielded no significant difference (44\% vs. 38\%, $\chi^2(3) = 0.73, p=.87$; see Table~\ref{table:prbmsexperiments}). Therefore, the difference between the learning behavior of people who seemed to use versus not use pseudo-rewards cannot be explained by situational factors. This suggests that those differences reflect inter-individual differences. 
\begin{table}[h!]
\centering
\small
\parbox{.45\linewidth}{
\begin{tabular}{ccc}
Model fam. & $r$ & $\phi$ \\ \hline
PR & 0.44 & 0.07 \\
No PR & 0.56 & 0.93 \\ \hline
\end{tabular}
}
\hfill
\parbox{.45\linewidth}{
\centering
\begin{tabular}{ccc}
Model fam. & $r$ & $\phi$ \\ \hline
PR & 0.38 & 0.02 \\
No PR & 0.62 & 0.98 \\ \hline
\end{tabular}
}
\caption{Family-level BMS analysis of pseudo-reward (PR) for Experiment 1 (left) and 2 (right)}
\label{table:prbmsexperiments}
\end{table}

Next to pseudo-reward, family-level BMS by grouping the models into two groups with and without the attribute suggest that models without hierarchical meta-control and without termination deliberation are preferred (see Table~\ref{table:hierarchicaltd}).
\begin{table}[h!]
\centering
\small
\parbox{.45\linewidth}{
\begin{tabular}{ccc}
Model fam. & $r$ & $\phi$ \\ \hline
HR & 0.15 & 0 \\
Non-HR & 0.85 & 1 \\ \hline
\end{tabular}
}
\hfill
\parbox{.45\linewidth}{
\centering
\begin{tabular}{ccc}
Model fam. & $r$ & $\phi$ \\ \hline
TD & 0.30 & 0\\
No TD & 0.70 & 1 \\ \hline
\end{tabular}
}
\caption{Family-level BMS comparing attributes of hierarchical meta-control (HR) and termination deliberation (TD)}
\label{table:hierarchicaltd}
\end{table}

\subsection{How well can our best models capture the qualitative changes in people's planning strategies?}
\begin{figure*}[h!]
\small
\begin{subfigure}{.33\linewidth}
  \centering
  \includegraphics[width=.99\linewidth, height=4cm]{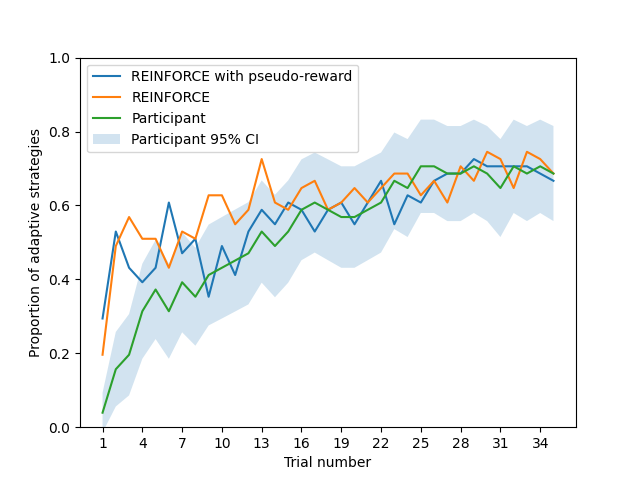}
  \caption{Far-sighted planning environment}
    \label{fig:increasing}
\end{subfigure}%
\begin{subfigure}{.33\linewidth}
  \centering
  \includegraphics[width=.99\linewidth, height=4cm]{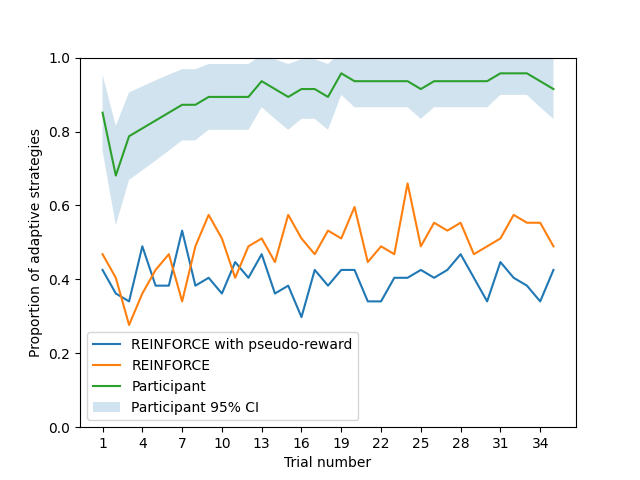}
  \caption{Short-sighted planning environment}
  \label{fig:decreasing}
\end{subfigure}%
\begin{subfigure}{.33\linewidth}
  \centering
  \includegraphics[width=.99\linewidth, height=4cm]{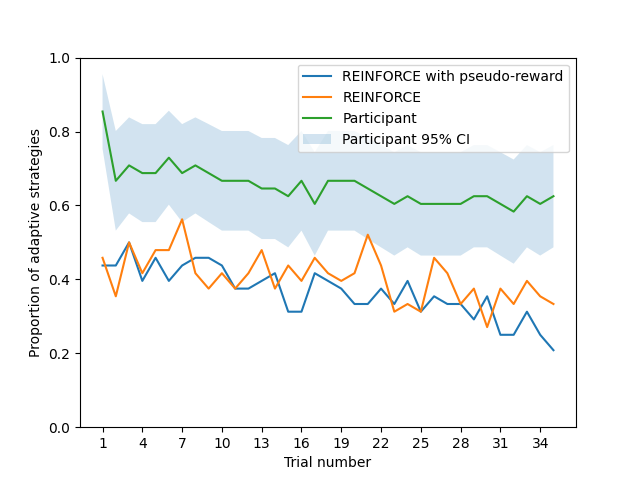}
  \caption{Best-first-search planning environment}
  \label{fig:constant}
\end{subfigure}
\caption{Average proportion of adaptive planning strategies in Experiment 1}
\label{fig:adaptiveproportion}
\end{figure*}
\begin{figure*}
\centering
\begin{minipage}{.3\textwidth}
  \centering
  \includegraphics[width=1.1\linewidth, height=4cm]{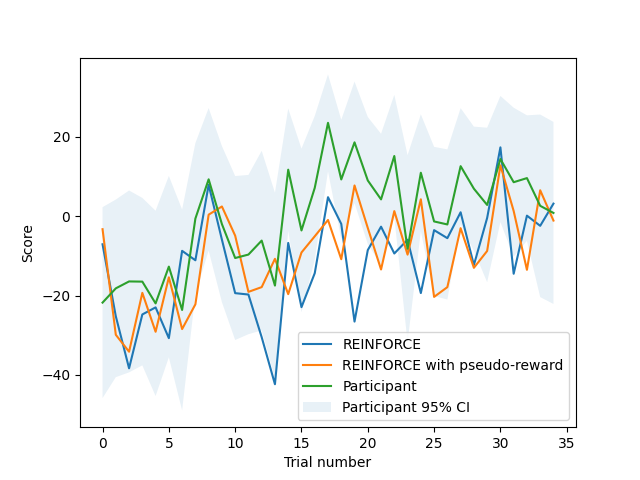}
  \centering
  \caption{Average score of the two best models and of the participants across 35 trials for both experiments}
  \label{fig:score}
\end{minipage}
\hfill
\begin{minipage}{.65\textwidth}
  \centering
\begin{subfigure}{.5\linewidth}
  \centering
  \includegraphics[width=.99\linewidth, height=4cm]{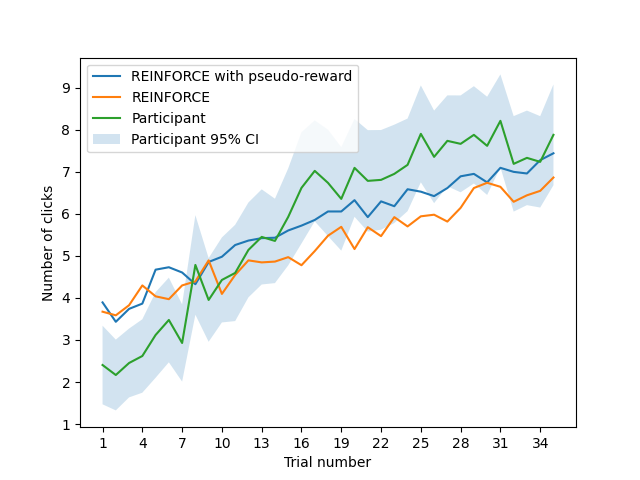}
  \caption{Planning is beneficial}
    \label{fig:highvariance}
\end{subfigure}%
\begin{subfigure}{.5\linewidth}
  \centering
\includegraphics[width=.99\linewidth, height=4cm]{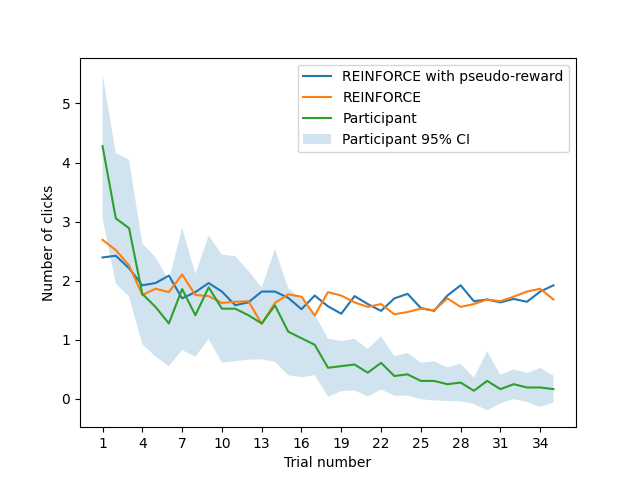}
  \caption{Planning is not beneficial}
  \label{fig:lowvariance}
\end{subfigure}
  \caption{Average number of clicks of the best two models and of the participants across 35 trials for Experiment 2}
  \label{fig:planningamount}
\end{minipage}
\end{figure*}
To examine, whether the two most promising models -- plain REINFORCE and REINFORCE with pseudo-reward -- can explain all phenomena observed in Experiments~1 and 2, we simulated participants' behavior in the three conditions of Experiment~1 and the two conditions of Experiment~2 with the fitted model parameters. Figure~\ref{fig:score} shows the increasing trend in the predicted level of resource-rationality over time across both experiments (Mann-Kendall test: all $S>214$ and $p<.01$ for both models and participants). This shows that our models can explain the observed increase in adaptiveness. Figure~\ref{fig:adaptiveproportion} shows the proportion of adaptive planning strategies in the first experiment. To determine whether a participant used an adaptive planning strategy on a given trial, we inspected the first click in each trial, which signals what kind of strategy has been used. First click on the farthest node signals the adaptive far-sighted strategy in the first condition; first click on an immediate node signals the near-sighted strategy in the second condition, and first click on the immediate and middle nodes signals the best-first-search in the third condition. 
Both models captured that people learned to increasingly more often rely on adaptive strategies in the condition, where far-sighted planning is beneficial (see Figure~\ref{fig:increasing}, Mann-Kendall test: increasing proportion of adaptive strategies for both models and participants; all $S>383, p<.01$). The plain REINFORCE model additionally captured the participants also learned to use increasingly more adaptive strategies in the environment that favored near-sighted planning (see Figure~\ref{fig:decreasing}; increasing trend for participants and plain REINFORCE: $S>227, p<.01$, no trend for REINFORCE with pseudo-rewards: $S=11, p=.89$). Moreover, both models captured that participants appeared to use increasingly fewer adaptive strategies in the environment that favored best-first-search planning (see Figure~\ref{fig:constant}; all $S<-217; p<.01$).\footnote{This might reflect shortcomings of the rule He, Jain, \& Lieder (2021) used to classify people's strategies in this environment.} Although the models captured these qualitative effects, from a quantitative perspective, they underutilized adaptive strategies in the environments where near-sighted planning and best-first-search planning are most beneficial but not in the environment that favored far-sighted planning. 
 
Both models can partly capture the participants' learning behavior in Experiment~2. For the conditions where planning is beneficial, both models correctly predicted that the amount of planning would increase significantly over time (see Figure~\ref{fig:highvariance}; Mann-Kendall test: all $S>516, p<.01$). For the condition where planning is less beneficial, the models predicted that the number of clicks would decrease to a nearly optimal level (see Figure~\ref{fig:lowvariance}, Mann-Kendall test: all $S<-167, p<.01$). Participants learned to decrease their amount of planning to an even greater extent and converged on planning less than the resource-rational strategy. This indicates that participants experience an additional cost that is not yet captured by our models. 

\section{Discussion and further work}
In this article, we tested 86 computational models of how people learn planning strategies against data collected in two experiments that tested different characteristics of metacognitive learning, namely the adaptation to different environment structures and the adaptation to different levels of planning costs. Overall, we found consistent evidence that the learning mechanism REINFORCE can largely capture the observed phenomena, like learning far-sighted planning strategies and adjusting the amount of planning. Moreover, we found that some people learn from self-generated pseudo-rewards for the value of information, whereas others do not. However, the REINFORCE models failed to learn short-sighted or best-first search planning strategies to the same extent as the participants.
%Simulations using the REINFORCE models suggest that it can explain how people learn far-sighted planning strategies and how people adopt their amount of planning. 
%However, the models failed to fully capture the learning behavior in environments, where short-sighted and best-first search planning are adaptive. 
This observation, combined with the high proportion of non-learning models, suggests that there is still room for improvement. %Moreover, different individuals might use different learning mechanisms. 
Furthermore, planning incurs cognitive costs above and beyond the cost of acquiring information \cite{He2022Where, Felso2020CogSci, callaway2022rational}. Therefore, further work can improve our models by incorporating these additional costs into the reward signals that the models learn from.

\bibliographystyle{apacite}

\setlength{\bibleftmargin}{.125in}
\setlength{\bibindent}{-\bibleftmargin}

\bibliography{CogSci_Template}

\end{document}